\documentclass[lettersize,journal]{IEEEtran}
\usepackage{amsmath,amsfonts}
\usepackage{algorithmic}
\usepackage{array}
\usepackage[caption=false,font=normalsize,labelfont=sf,textfont=sf]{subfig}
\usepackage{textcomp}
\usepackage{stfloats}
\usepackage{url}
\usepackage{verbatim}
\usepackage{graphicx}
\usepackage{xcolor}
\usepackage{float}
\usepackage{multirow}
\usepackage{adjustbox}

\def\BibTeX{{\rm B\kern-.05em{\sc i\kern-.025em b}\kern-.08em
    T\kern-.1667em\lower.7ex\hbox{E}\kern-.125emX}}
\usepackage{balance}

\DeclareMathOperator*{\argmin}{arg\,min}

\begin{document}

\title{Land Surface Temperature Super-Resolution with a Scale-Invariance-Free Neural Approach: Application to MODIS}

\author{
\IEEEauthorblockN{Romuald Ait-Bachir\IEEEauthorrefmark{1}\IEEEauthorrefmark{2}, Carlos Granero-Belinchon\IEEEauthorrefmark{1}\IEEEauthorrefmark{2}, Aur\'elie Michel\IEEEauthorrefmark{3}, Julien Michel\IEEEauthorrefmark{4}, Xavier Briottet\IEEEauthorrefmark{3} and Lucas Drumetz\IEEEauthorrefmark{1}\IEEEauthorrefmark{2}} \\
\IEEEauthorblockA{\IEEEauthorrefmark{1}IMT Atlantique, Lab-STICC, UMR 6285, 29238, CNRS, Brest, France} \\
\IEEEauthorblockA{\IEEEauthorrefmark{2}ODYSSEY Team-Project, INRIA Ifremer IMT-Atl., 35042, CNRS, Brest, France} \\
\IEEEauthorblockA{\IEEEauthorrefmark{3}ONERA-DOTA, University of Toulouse, F-31055 Toulouse, France} \\
\IEEEauthorblockA{\IEEEauthorrefmark{4} CESBIO, University de Toulouse, CNES, CNRS, INRAE, IRD, UT3, 31401 Toulouse, France.}  \thanks{corresponding author: carlos.granero-belinchon@imt-atlantique.fr}\thanks{This research was funded by C.N.E.S in the A.P.R CES ``Th\'eia Temp\'erature de Surface et \'Emissivit\'e'' (2022-2025) framework.}}

\markboth{Journal of \LaTeX\ Class Files,~Vol.~18, No.~9, September~2020}%
{How to Use the IEEEtran \LaTeX \ Templates}

\maketitle

\begin{abstract}
%150-250 words
Due to the trade-off between the temporal and spatial resolution of thermal spaceborne sensors, super-resolution methods have been developed to provide fine-scale Land Surface Temperature (LST) maps. Most of them are trained at low resolution but applied at fine resolution, and so they require a scale-invariance hypothesis that is not always adapted. The main contribution of this work is the introduction of a Scale-Invariance-Free approach for training Neural Network (NN) models, and the implementation of two NN models, called Scale-Invariance-Free Convolutional Neural Network for Super-Resolution (SIF-CNN-SR) for the super-resolution of MODIS LST products. The Scale-Invariance-Free approach consists on  training the models in order to provide LST maps at high spatial resolution that recover the initial LST when they are degraded at low resolution and that contain fine-scale textures informed by the high resolution NDVI. The second contribution of this work is the release of a test database with ASTER LST images concomitant with MODIS ones that can be used for evaluation of super-resolution algorithms. We compare the two proposed models, SIF-CNN-SR1 and SIF-CNN-SR2, with four state-of-the-art methods, Bicubic, DMS, ATPRK, Tsharp, and a CNN sharing the same architecture as SIF-CNN-SR but trained under the scale-invariance hypothesis. We show that SIF-CNN-SR1 outperforms the state-of-the-art methods and the other two CNN models as evaluated with LPIPS and Fourier space metrics focusing on the analysis of textures. These results and the available ASTER-MODIS database for evaluation are promising for future studies on super-resolution of LST.
\end{abstract}

\begin{IEEEkeywords}
Super-Resolution, LST, Neural Networks, MODIS, ASTER 
\end{IEEEkeywords}

%\tableofcontents

\section{Introduction}

\subsection{Overview}

Land Surface Temperature (LST) is one of the Essential Climate Variables (ECVs) to describe the changing climate of the Earth~\cite{Bojinski2014}. It refers to the temperature of the Earth's surface and is a key variable in the physics of the land surface and its interactions with the atmosphere. It is used for a wide range of applications, such as the detection and characterization of urban heat islands (UHI) and heatwaves~\cite{Almeida2021, Albright2011, Dousset2011,Yang2024}, droughts~\cite{Wan2004, Rahmi2021,Orimoloye2021}, forest fires~\cite{Kumar2022, Zhao2024}, inland water bodies~\cite{Sobrino2024}, or for the analysis of warming trends in different parts of the world~\cite{Liu2021, Wang2022, Waring2023} among others~\cite{Guillevic2018}. It also allows to estimate the surface energy balance~\cite{anderson-2008-therm-based} and the evapo-transpiration of the vegetation~\cite{anderson_use_2012}. LST is retrieved from two different kinds of sensors: those using Passive MicroWave (PMW) wavelengths and those using Thermal InfraRed (TIR) ones. On the one hand, PMW sensors acquire for all weather conditions, even through rain or clouds, but current PMW-based sensors present very low spatial resolutions of the order of 10-25 km~\cite{Duan2020}. On the other hand, commonly used TIR sensors require cloud-free conditions but present better spatial resolutions of 70 m to 1 km~\cite{Li2023}. Both PMW and TIR sensors provide information at regional or global scales, but only TIR sensors provide LST maps at a sufficient spatial resolution for monitoring and studying local phenomena \cite{Anderson2021, Yang2021}. Although TIR sensors provide finer spatial resolution maps, there is a trade-off between spatial resolution and temporal one: while high spatial resolution sensors with resolutions of around 100~m present very low acquisition frequencies of the order of one image per month or less (due to clouds), lower spatial resolution sensors such as MODIS, with resolutions of the order of 1 km acquire around two images per day. MODIS provides one of the most widely used LST products because it spans more than two decades~\cite{Phan2018, Reiners2023}. As LST displays a large heterogeneity in both space and time, monitoring the LST at both high temporal and spatial resolution is needed for a significant number of applications within the land and solid Earth, health and hazards and security and surveillance frameworks, as highlighted by~\cite{Sobrino2016}. For example, studies related to public health, volcanology, or urban climatology would benefit from LST series with a higher acquisition frequency~\cite{Liu2012, Mitraka2019, Ramsey2020}. Thus, with the objective of providing LST maps with both high spatial and temporal resolutions, there is a growing literature in super-resolution methods aimed at increasing the spatial resolution of the LST images, notably for MODIS~\cite{Li2019, Yoo2020}. Although there are studies on improving the spatial resolution of PMW-based LST maps~\cite{Favrichon2021, Zhong2021, Li2024}, this field is beyond the scope of this paper that aims to address the trade-off of TIR sensors.

\subsection{Related Works}\label{sec:relatedworks}

Super-resolution of Land Surface Temperature images is commonly done with statistical models based on empirical relationships between LST and land features estimated from Visible and Near InfraRed (VNIR) satellite images. This approach was initially justified by the inverse relationship between Normalized Difference Vegetation Index (NDVI) and LST~\cite{Cai2018, Govil2019}. Later, several studies showed that VNIR indices are adapted for LST super-resolution~\cite{Kumar2015,Ferreira2019,GraneroBelinchon2019}. The most commonly used statistical models for LST super-resolution are based on simple linear regressions between the LST and VNIR features. Some of these models are Distrad, TsHARP, ATPRK or AATPRK~\cite{Essa2012,  Kustas2003, TsHARP2nd, Wang2015, Wang2016}. Granero-Belinchon et al. 2019 illustrated that these models are only slightly dependent on the VNIR feature used in the regression and that ATPRK (Area-To-Point Regression Kriging) and AATPRK (Adaptive-Area-To-Point-Regression-Kriging) with a Kriging interpolation for the residuals' definition slightly outperform TsHARP and DisTrad~\cite{GraneroBelinchon2019}. Recently, more complex models such as Data Mining Sharpener (DMS) using regression trees appeared~\cite{Gao2012, Xue2020, Guzinski2019}. All these models use relationships between the LST and VNIR features that are optimized at the LST coarse resolution and applied at the high resolution of the VNIR feature. So, these models use a reduced scale training approach.

In the recent years, Deep Learning has been widely used for super-resolution applications of remote sensing images in the VNIR, where a lot of different approaches have been developed: Convolutional Neural Networks (CNN)~\cite{Gargiulo2019, Lanaras2018}, Generative Adversarial Networks~\cite{Brodu2017} or diffusion models~\cite{Xiao2023} among others. However, Deep Learning has been only scarcely used for LST super-resolution~\cite{Nguyen2022,Chen2024,Choe2017}. A CNN has been used on images from MODIS~\cite{Nguyen2022} to increase the spatial resolution by a factor of four. Multilayer perceptrons have also been used to downscale MODIS LST to Landsat spatial resolution (by a factor of ten)~\cite{Choe2017} and more recently, \cite{Chen2024} proposed a diffusion model to downscale Landsat LST by factors of four and eight. However, in the field of remote sensing, a frequent difficulty in super-resolution is the absence of a high resolution reference image to be used in the training step. This is particularly true for LST super-resolution. So, in order to apply supervised learning schemes, models are commonly trained at reduced scale, \textit{i.e.} trained at degraded lower resolutions using the initial images as a reference to later apply these trained models at higher resolutions~\cite{Nguyen2022, Chen2024, Choe2017}.

All the aforementioned statistical and deep learning models use a scale-invariance hypothesis: models optimized at coarse resolution can be used at high resolution. This hypothesis can be problematic depending on the heterogeneity of the studied surface or the range of scales studied.

\subsection{Motivation and contributions}

Several studies demonstrated that this scale-invariance hypothesis can lead to bad resolved small scale textures~\cite{Nguyen2022,GraneroBelinchon2019,Merlin2010}. Changes in the statistical laws learned at different resolutions~\cite{GraneroBelinchon2019} lead to an important loss in performance of the models when applied to resolutions higher than the one of training~\cite{Nguyen2022}. To overcome this limitation, this article presents a new deep learning model that is trained at full scale, \textit{i.e.} trained without scale-invariance hypothesis. The optimization problem statement of this model is based on a variational formulation~\cite{Barker2004,Melinc2024}.

The main contributions of this research are:
\begin{itemize}
    \item A Scale-Invariance-Free Convolutional Neural Network for super-resolution (SIF-CNN-SR). This model increases the spatial resolution of a given low resolution LST by imposing that 1) the initial low resolution LST image is recovered when the spatial resolution is degraded, and 2) the high resolution LST presents small scale textures from the NDVI. These two constraints correspond respectively to consistency and synthesis properties~\cite{Palsson2016}. This approach aims at reconstructing the high frequencies with better fidelity.
    \item The demonstration that the SIF-CNN-SR approach presents a performance at the state of the art level while greatly outperforming some commonly used super-resolution approaches such as TsHARP and ATPRK.
    %\item The limitations of the local modelization used the approach to make the texture appear. (ie some variations in the NDVI do not mimic the LST ones leading to over or underestimations).
    \item A super-resolution evaluation dataset associating concomitant coarse resolution MODIS LST images and high resolution ASTER LST ones which allows to evaluate LST super-resolution algorithms on non simulated data to quantify their performance. 
\end{itemize}

\section{Scale-Invariance-Free Convolutional Neural Network for super-resolution}

This section presents the proposed model for LST super-resolution, as well as the NN architecture.

\subsection{Model}\label{sec:model}

Our model is inspired by classical variational schemes:

\begin{equation}\label{eq:3dvar}
    D(\theta) = \argmin_{\theta} (|| y - H\left(x(\theta)\right)|| + \lambda R(\theta))
\end{equation}

\noindent where $x$ is the reconstructed state of the system, $y$ is an observation of the system state, $H$ is an operator modeling the observation $y$ from the state $x$, $R$ is a regularization term used to impose desired properties to the reconstructed state $x$, $\lambda$ is a constant weighting the significance of the regularization term and $|| \,||$ is a given norm. Both $x$ and $R$ depend on $\theta$, the parameters to be optimized. We can cast variational approaches as an optimization problem with two criteria: being able to obtain a state of the system $x$ that is close to the observation $y$ and matching the properties imposed by the regularization term.

We propose a Neural Network model (SIF-CNN-SR) to downscale a LST image observed at low spatial resolution $T^{(l)}_{obs}$ to a higher spatial resolution $T^{(h)}_{sr}$. The superscripts $(l)$ and $(h)$ indicates respectively low and high resolution products. The model is trained in a self-supervised way and takes as input a couple of (LST, NDVI) images of the same region acquired at the same time. The input LST correspond to the initial observation at low resolution $T^{(l)}_{obs}$ while the input NDVI is acquired at high resolution $V^{(h)}_{obs}$. The model is trained in order to provide LST at high resolution that 1) when degraded to low spatial resolution recovers the initial LST used as input and 2) contains fine-scale textures informed by the high resolution NDVI used as input. Making the parallelism with variational approaches, these two objectives correspond respectively to provide a super-resolution LST close to the observed one, and close to a ``background state'' defined by the texture of the NDVI.

In order to ensure both conditions, the optimization problem is written as follows:

%\begin{align}
%\hat{\theta} = & \argmin_{\theta} \,  \alpha \left( \lvert\lvert \gamma G(V^{(h)}_{obs}) - G(\Psi_{\theta}(V^{(h)}_{obs},T^{(l)}_{obs})) \rvert\rvert \right) +  \label{eq:optimization} \\
%&+ (1-\alpha) \left( \lvert\lvert T^{(l)}_{obs} - H(\Psi_{\theta}(V^{(h)}_{obs},T^{(l)}_{obs})) \rvert\rvert \right) \nonumber \\
%\hat{T}^{(h)}_{sr} = & \,  \Psi_{\hat{\theta}}(V^{(h)}_{obs},T^{(l)}_{obs})
%\end{align}

\begin{align}
\hat{\theta} = & \argmin_{\theta} \,  \alpha \left[ J \left(\gamma G(V^{(h)}_{obs}),G(\Psi_{\theta}(V^{(h)}_{obs},T^{(l)}_{obs})) \right) \right] +  \nonumber \\
&+ (1-\alpha) \left[ J \left(T^{(l)}_{obs}, H(\Psi_{\theta}(V^{(h)}_{obs},T^{(l)}_{obs}))\right) \right] \label{eq:optimization}
\end{align}

\noindent where $\Psi_{\theta}$ is the model providing high-resolution LST, $\theta$ are the parameters of the model to be learned, $G$ is an operator defining small scale textures and $H$ is an observation operator modeling the degradation of spatial resolution from high to low resolution. $\gamma$ is a scaling coefficient to adapt the amplitude of the textures of NDVI and LST in the comparison. Finally, $\alpha$ is a weight coefficient used to compensate the significance of each term in the optimization and $J(a,b)$ is a given discrepancy measure comparing $a$ and $b$. The LST at high spatial resolution provided by the model with optimized parameters $\hat{\theta}$ is:

\begin{equation}
T^{(h)}_{sr} = \,  \Psi_{\hat{\theta}}(V^{(h)}_{obs},T^{(l)}_{obs})
\end{equation}

%Because of lack of information and for simplicity, we decided to choose $B^{-1}=I$ and $R^{-1}=I$.
%Equation~\ref{eq:3dvar} is a particular case of equation~\ref{eq:optimization} where $J$ is the square of the inverse of the covariance matrix $B$ and $R$, with $B=R$.

We identify two terms in (\ref{eq:optimization}): a reconstruction term and a texture one.

\begin{align}
\mathcal{L}_{rec} &= \left[ J \left( T^{(l)}_{obs}, H(\Psi_{\theta}(V^{(h)}_{obs},T^{(l)}_{obs})) \right) \right] \\
\mathcal{L}_{texture} &= \left[ J \left(\gamma G(V^{(h)}_{obs}),G(\Psi_{\theta}(V^{(h)}_{obs},T^{(l)}_{obs})) \right) \right]
\end{align}

The reconstruction term $\mathcal{L}_{rec}$ is used to obtain physical values of the LST since it seeks to recover the observed LST ($T^{(l)}_{obs}$) from the super-resolution LST ($T^{(h)}_{sr}$) by degrading the spatial resolution of the latter. Consequently, the observation operator $H$ consists in a low pass filtering of $T^{(h)}_{sr}$ simulating the effect of the thermal sensor Modulation Transfer Function (MTF)~\cite{Xiong2020}. This low pass filtering is done by convolving with a Gaussian kernel $K$. Then, a bicubic interpolation is used to reduce the size of the image to match the size of $T^{(l)}_{obs}$.

The texture term $\mathcal{L}_{texture}$ is inspired by the style-transfer defined by \cite{Johnson2016}. It is used to make the model add small scale textures that are present in $V^{(h)}_{obs}$ into $T^{(h)}_{sr}$ \textit{i.e.} to transfer the textures from the NDVI into the LST. Two definitions of the small scale texture operator $G$ are used:

\begin{itemize}
\item The first one combines the gradient with two additional diagonal derivatives. We write it $G(I)=\nabla_D I$ and is computed in practice by convolving any image $I$ with four derivative Sobel filters of size $3\times3$ following respectively horizontal, vertical and diagonal directions.

\item The second one consists in a high-pass filtering, done in practice by:
\begin{equation}\label{eq:highpass}
    G(I) = I - K \ast I
\end{equation}

\noindent with $K$ the Gaussian kernel modeling the MTF of the thermal sensor. So, the high-frequency content of any image $I$ is defined as the difference between the image itself and a low-pass filtered version $K \ast I$.
\end{itemize}

\subsection{Architecture}\label{sec:arch}

The proposed SIF-CNN-SR model is not constrained by a specific architecture for $\Psi_{\theta}$. In this work, $\Psi_{\theta}$ is a U-Net~\cite{Ronneberger2015}, which is a fully convolutional neural network with an encoder-decoder architecture and long-skip connections between the encoder and the decoder at each level. More particularly, our model grounds on the Multi-Residual U-Net used in Nguyen et al.~\cite{Nguyen2022} to downscale the LST of MODIS. U-Nets are commonly used for super-resolution applications~\cite{Chen2024a,Kalluvila2023,Hu2019}.

\begin{figure*}[h]
    \centering
    \includegraphics[width=\linewidth]{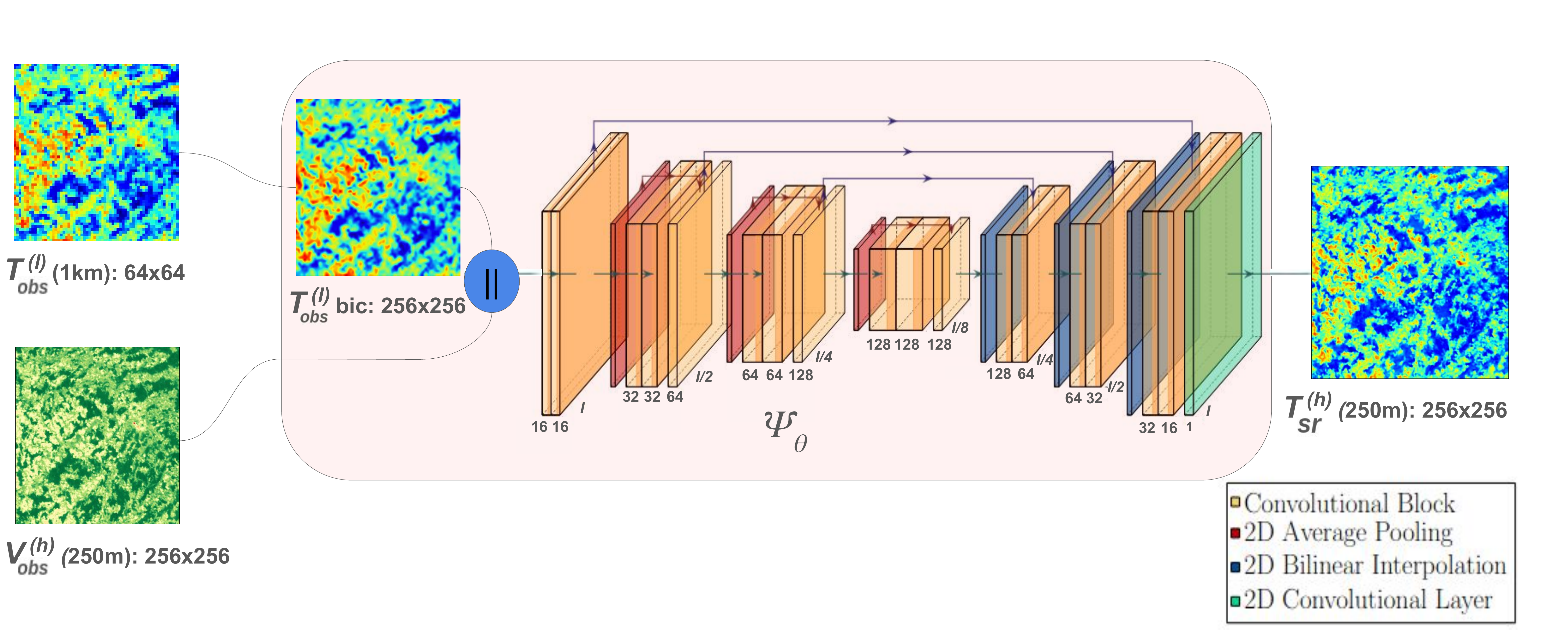}
    \caption{Schematic representation of the U-net architecture of $\Psi_{\theta}$. Orange blocks correspond to convolutional blocks with (Conv2d, Batchnorm2d, ReLU), red blocks are average pooling reducing the size of the image by a factor two along each dimension, blue blocks are bilinear interpolation increasing the size of the image by a factor two and the green block is a two dimensional convolutional layer. The number of channels of each two dimensional convolutional layer is indicated in the figure. The input is the concatenation ($||$) along the channel dimension of the high resolution NDVI $V^{(h)}_{obs}$ and the bicubic interpolation of $T^{(l)}_{obs}$. The output is the super-resolution $T^{(h)}_{sr}$ at the spatial resoluion of $V^{(h)}_{obs}$.}\label{fig:archi}
\end{figure*}

Our multi-residual U-net takes as input the concatenation ($||$) along the channel dimension of $V_{obs}^{(h)}$ and the bicubic interpolation of $T_{obs}^{(l)}$ and is made of an encoder and its symmetrical decoder, each one with 3 levels, see figure~\ref{fig:archi}. The model first present two consecutive convolutional blocks (Conv2d, Batchnorm2d, ReLU) and then the three levels of the encoder. Each level of the encoder consists of an average pooling reducing the size of the image by two on each dimension, followed by three convolutional blocks with a residual connection between the input of the first block and the output of the second one. The decoder has also three levels, each one with a non-trainable bilinear interpolation increasing the size of the image by two, followed by two convolutional blocks. Finally the output of the last convolutional block of the decoder is passed through a 2d convolutional layer. The corresponding stages of the encoder and the decoder are linked with concatenated long skip connections. All the 2d convolutional layers have kernel size $3$, stride $1$, padding $1$ in replicate mode, and null bias. The total number of trainable parameters is 417 009~\footnote{The trained models are available at https://github.com/cgranerob/Land-Surface-Temperature-Super-Resolution-with-a-Scale-Invariance-Free-Neural-Approach.}. %$\Psi_{\theta}$ takes as inputs $V^{(h)}_{obs}$ and $T^{(l)}_{obs}$ and provides as output $T^{(h)}_{sr}$.

\section{Application to MODIS LST super-resolution}

In this work, we use SIF-CNN-SR to downscale MODIS LST from 1 km of spatial resolution to 250 m, that is the spatial resolution of the NDVI from MODIS. The model takes as input a couple of (LST, NDVI) MODIS images of the same region acquired at the same time. The model is trained in order to provide MODIS LST at 250 m that 1) when degraded into 1 km of spatial resolution recovers the initial MODIS LST and 2) contains fine-scale textures informed by the MODIS NDVI at 250 m used as input. For validation, we use ASTER data. Both sensors are described in the following.

\subsection{Data}

This section introduces the specifications of the Moderate-Resolution Imaging Spectroradiometer (MODIS) and the Advanced Spaceborne Thermal Emission and Reflection Radiometer (ASTER) sensors and their products. Then, it presents the construction of the training, validation, and test datasets used in this study.

\subsubsection{MODIS}

MODIS is a sensor that acquires data in 36 spectral bands from the visible to the thermal infrared domains. MODIS is onboard the TERRA and AQUA platforms allowing the observation of the entire surface of the Earth with a revisit time of four times per day considering both satellites. Each MODIS image covers an area of 1200x1200 km$^2$.

In this work, we focus on two MODIS products: MOD21A1 which provides Land Surface Temperature at 1 km of spatial resolution and MOD09GQ which provides reflectances in the Red and NIR domains at 250 m of spatial resolution \cite{MODISLST, MODISNDVI}. Both products provide images of the same land surfaces and acquired at the same acquisition time. The LST MOD21A1 product results from the application of the Temperature and Emissivity Separation (TES) algorithm~\cite{Gillespie1998} to the radiance of the three MODIS bands in the TIR: bands 29, 31 and 32. MOD21A1D only contains LST images acquired during daytime from the MODIS sensor installed on the TERRA satellite. We combine the NIR and Red reflectances from MOD09GQ to generate NDVI images. 

\subsubsection{ASTER}
ASTER is also a sensor onboard the TERRA platform. It acquires data over 14 spectral bands going from the VNIR to the TIR spectral domains. ASTER provides LST images at 90 m of spatial resolution, and a revisit time of 16 days. 
%The images cover land surfaces of 60x60 km$^2$. 
In this work, we use the ASTER AST\_08 product which provides LST retrieved from the application of the TES algorithm to the 5 TIR bands of ASTER \cite{ASTER}.

\subsection{Training, validation and evaluation datasets}

\subsubsection{Training and validation dataset}\label{sec:datatrain}

\begin{figure}[htbp]
    \centering
    \includegraphics[width=\linewidth]{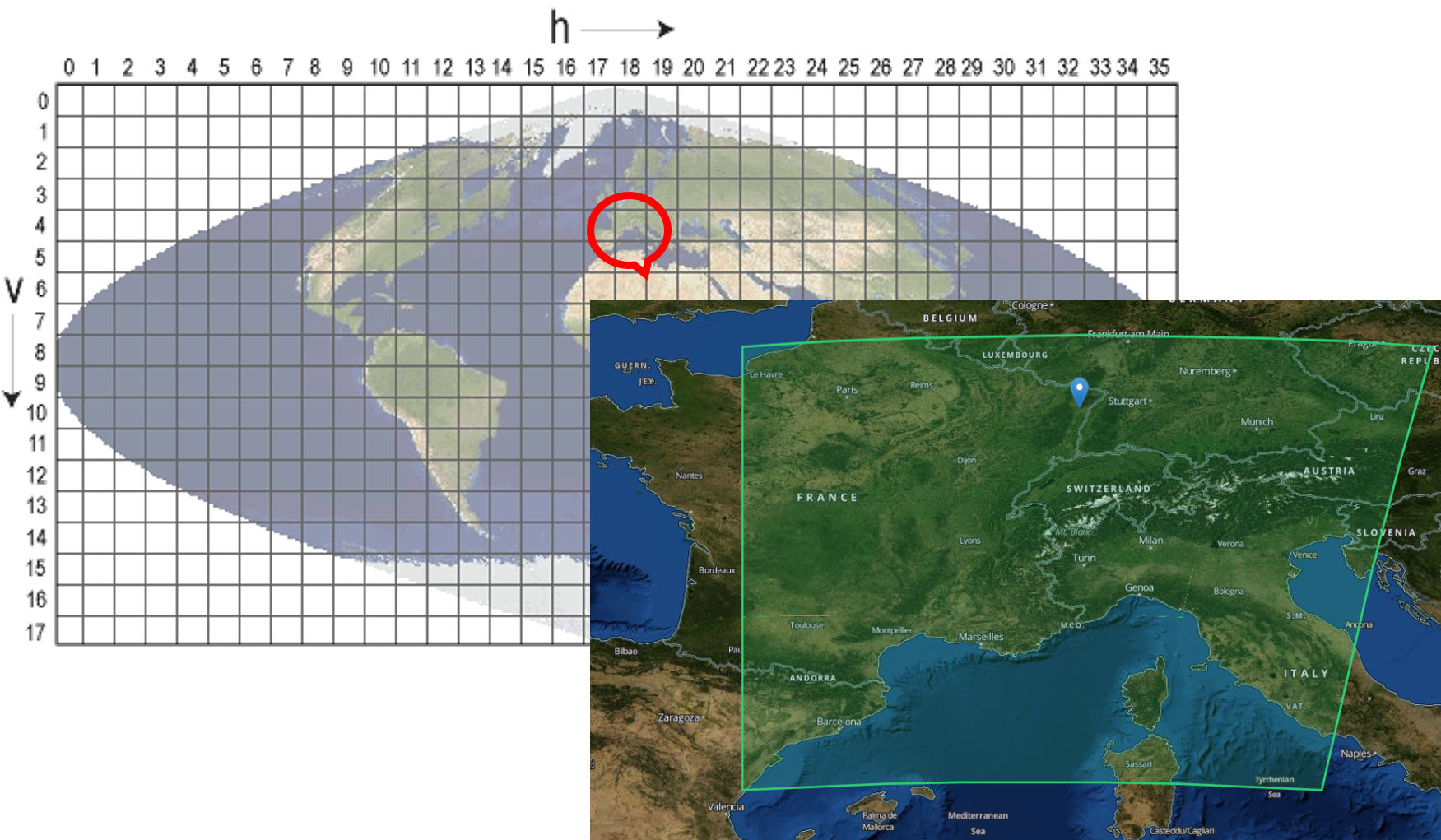}
    \caption{Center Europe area (h18v04 MODIS tile) used in this study.}
    \label{fig:tile}
\end{figure}

The training and validation dataset contains LST and NDVI concomitant images coming from the MOD21A1D and MOD09GQ MODIS products, see table~\ref{tab:data}. It spans over three years, from January 1st 2017 until December 31st 2019. The used images (tile h18v04) cover the center of Europe, mainly France, Germany, Italy, Austria and Switzerland, see figure~\ref{fig:tile}. In total, 1095 pairs of MODIS (LST, NDVI) images covering areas of $1200 \times 1200$ km$^2$ are used. No nighttime images were considered to avoid any registration error between daytime NDVI and nighttime LST as well as to avoid different thermal dynamics between day and night that can complicate the learning process.

\begin{table*}
 \caption{Summary information of training, validation and evaluation datasets.}\label{tab:data}
\begin{tabular}{ |p{2cm}|p{2.5cm}|p{2cm}|p{2cm}|p{2cm}|p{1.5cm}|p{1.5cm}|p{1.5cm}|}
 \hline
Sensor              & Product & Region        & Time period        & Spatial Resolution  & Images for training & Images for validation & Images for evaluation \\
\hline
\multirow{2}{*}{MODIS} & MOD21A1D (LST) & \multirow{5}{*}{Center Europe} & \multirow{2}{*}{2017-2019} & 1km  & \multirow{2}{*}{4900} & \multirow{2}{*}{3333} & \multirow{2}{*}{0} \\ \cline{2-2} \cline{5-5}
                      & MOD09GQ (NDVI)&  &  & 250m &  &  &  \\ \cline{1-2} \cline{4-8}
\multirow{2}{*}{MODIS} & MOD21A1D (LST) &  & \multirow{3}{*}{2020-2023} & 1km  & \multirow{3}{*}{0}    & \multirow{3}{*}{0}    & \multirow{3}{*}{79} \\ \cline{2-2} \cline{5-5}
                    & MOD09GQ (NDVI)&  &  & 250m &     &     &  \\ \cline{1-2} \cline{5-5}
ASTER               & AST\_08 (LST) &  &  & 90m  &     &     &  \\
 \hline
\end{tabular} 
\end{table*}

The data is further pre-processed by slicing the images into smaller patches covering a land surface of $64 \times 64$ km$^2$ ($64 \times 64$ and $256 \times 256$ pixels for low resolution and high resolution products respectively) without clouds neither pixels evaluated by the NASA as low quality. In total, the training and validation dataset is made up of 8233 pairs of patches. It is then split following a 0.6/0.4 ratio leading to a training dataset containing 4900 pairs of images and a validation dataset containing 3333 pairs of images. During training, each LST and NDVI image is centered and standardized using the mean and variance of the dataset.

\subsubsection{Evaluation dataset} \label{sec:evaldata}

The dataset for evaluation contains concomitant MODIS and ASTER images covering most of the area of the MODIS tile h18v04 \textit{i.e.} the same region used in the training. These images are acquired during three years starting from January 1st 2020 (so without overlapping with the training dataset). For each date, the corresponding MODIS data (MOD21A1D and MOD09GQ) is associated to the AST\_08 data, see table~\ref{tab:data}. Concomitant MODIS and ASTER LST acquisitions are needed because of the fast dynamics of the temperature that can lead to strong LST evolutions in times of the order of 1 hour~\cite{Lu2021}. 

In order to pair ASTER and MODIS images several pre-processing steps are done. First, the ASTER spatial resolution is degraded from 90 m to 250 m (the spatial resolution of the NDVI image of MODIS) by convolving it with a Gaussian kernel with half-kernel width of $250$ m. Second, they are both projected into the same UTM (Universal Transverse Mercator) coordinate system and co-registered. Third, since ASTER images cover smaller areas, the MODIS image is cropped to cover the same region as the ASTER one. The evaluation dataset contains 79 pairs of concomitant LST images from ASTER and (LST, NDVI) couples from MODIS of 64 $\times$ 64 km$^2$.

Even if ASTER and MODIS are both onboard the TERRA platform, small errors in co-registration and small differences in acquisition times can still exist. In particular, for the couples defining our evaluation dataset the mean difference in acquisition times between ASTER and MODIS is $1.5$ minutes with a variance of $4.2$ minutes. The maximum and minimum time differences observed are respectively of $15.1$ and $1.2$ minutes. 
%
%For the couples defining our second test dataset the mean difference in acquisition times between ASTER and MODIS is $2.9$ minutes with a variance of $3.2$ minutes. The maximum and minimum time differences observed are respectively of $11.6$ and $0.7$ minutes. We can then consider that these differences in the acquisition times are negligible compared to the times needed for the LST to vary significantly.

\subsection{Training procedure}

The model $\Psi_{\theta}$ takes as inputs a couple of MODIS (LST, NDVI) images. In order for both images to have the same dimensions and observe the same area the MODIS LST is downscaled using bicubic interpolation. The model $\Psi_{\theta}$ provides a super-resolution LST image.

The loss function optimized during the training procedure is defined in (\ref{eq:optimization}) with $J$ being the Huber loss for this specific application since it is specially robust to outliers~\cite{Goshin2021}. 

The choice of the loss hyperparameters, $\alpha$ and $\gamma$, is critical. On the one hand, $\alpha \in [0,1]$ weights the reconstruction and texture terms of the loss and should be chosen to equilibrate the significance of each term in the minimization. On the other hand, $\gamma$ is a scale factor used to compensate the difference of the range of variations between the gradients of NDVI and LST. This difference is still present after standardization. $\gamma$ is usually negative due to the inverse relationship between the LST and the NDVI~\cite{Cai2018,Govil2019}.

In this study, two models with slightly different definitions of the small scale texture $G$ are trained: SIF-CNN-SR1 and SIF-CNN-SR2.
SIF-CNN-SR1 uses Sobel filters to define the small scales, see section~\ref{sec:model}. The choice of $\gamma=-0.5$ was inferred from a small grid search on the values $\left\lbrace -0.1, -0.25, -0.5, -0.75, -1\right\rbrace$. $\alpha$ was set to 0.99. We expected a much higher texture loss than the reconstruction one, explaining the choice of $\alpha=0.99$ to compensate.
SIF-CNN-SR2 uses high-pass filtering to define the small scales, see section~\ref{sec:model}. The choice of $\gamma=-0.25$ was inferred from a grid search on $\left\lbrace -0.1, -0.25, -0.5, -0.75, -1\right\rbrace$. $\alpha$ was set to $0.1$ due to the alternative representation of the texture term leading to much lower texture loss values.  

The learning rate was set to 0.0001 and the batch size to 32. The model was trained for 200 epochs. The training was made using Python 3.9.19 and Pytorch 2.3.0. It used a Quadro RTX 8000 with 46 GO of VRAM. Table \ref{tab:synthesis} synthesizes the main differences between SIF-CNN-SR1 and SIF-CNN-SR2.

\begin{table}[ht]
    \centering
    \caption{Synthesis of the parameters for models SIF-CNN-SR1 and SIF-CNN-SR2 both trained during 200 epochs with a learning rate of 0.0001 and batch size of 32. The loss function used is defined in~(\ref{eq:optimization}) with $H$ defined as the MTF of MODIS.}
    \begin{tabular}{cccc}
        Model       & $\alpha$ & $\gamma$ & G \\
        \hline
        SIF-CNN-SR1 & 0.99     &  -0.5    & Sobel filter     \\
        \hline
        SIF-CNN-SR2 & 0.10     &  -0.25   & High pass filter \\
        \hline
        \\
    \end{tabular}
    \label{tab:synthesis}
\end{table}

%\begin{table}[ht]
%    \centering
%    \begin{adjustbox}{width={\linewidth}}
    %\small
%    \caption{Synthesis of the parameters for each model SIF-CNN-SR1 and SIF-CNN-SR2 both trained during 200 epochs with a learning rate of 0.001 and batch size of 32. The loss function used is defined in~\ref{eq:optimization} with $H$ defined with MTF of MODIS.}
%    \begin{tabular}{ccccc}
%        Parameters & Epochs & Learning & Batch & $\alpha$ \\
%        & Number & Rate & Size &  \\
%        \hline
%        SIF-CNN-SR1 & 200 & 0.0001 & 32 & 0.99\\
%        \hline
%        SIF-CNN-SR2 & 200 & 0.0001 & 32 & 0.1 \\
%        \hline
%        \\
%        Parameters & $\gamma$ & $G$ & $H$ & Function \\
%        & & & &  Loss \\
%        \hline
%        SIF-CNN-SR1 & -0.5 & Sobel  & MTF & (\ref{eq:optimization}) \\
%        & & filter & & \\
%        \hline
%        SIF-CNN-SR2 & -0.25 & High & MTF & (\ref{eq:optimization})\\
%        & &  frequency & & \\
%        & & filter & &\\
%        \hline
 %       \\
%    \end{tabular}
%    \label{tab:synthesis}
%    \end{adjustbox}
%\end{table}

\subsection{Benchmarks}\label{sec:benchmarks}

\subsubsection{State-of-the-art statistical methods for super-resolution}

Four common methods from the state of the art are used for benchmarking SIF-CNN-SR1 and SIF-CNN-SR2: Bicubic interpolation, TsHARP, ATPRK and DMS. While bicubic interpolation is just a two dimensional interpolation of the image based on polynomials, TsHARP, ATPRK and DMS are sharpening techniques specifically developed for remote sensing applications and ground on a established statistical relationship between LST and variables obtained from the VNIR domain, such as the NDVI we use here.

TsHARP, ATPRK and DMS share a common strategy for sharpening:

\begin{itemize}
\item First, they consider a given relationship between LST and NDVI at low resolution (the initial resolution of the LST).

\begin{equation}
    T^{(l)}_{obs} = f(V^{(l)}_{obs}) + \Delta^{(l)} T
\end{equation}

\noindent with $\Delta^{(l)} T$ the residuals between the model and the ground-truth at low resolution. In the case of TsHARP and ATPRK, $f$ is a linear function, and the slope and intercept of the linear relationship are obtained by least squares minimization. In the case of DMS, $f$ is defined by m5 regression trees~\cite{Quinlan1992} with optimized parameters. DMS is much more complex than TsHARP and ATPRK by being a piece-wise linear model due to the leaves of the m5 tree being linear regressions. It is based on multiple locally and a globally fitted bagged regression trees which link non-linearly a NDVI to a LST. All these methods define the low resolution residuals $\Delta^{(l)} T$ for each pixel as $\Delta^{(l)} T = T^{(l)}_{obs} - f(V^{(l)}_{obs})$. 

Straightforward generalizations of TsHARP and ATPRK, replacing the linear $f$ by a bilinear one, and the current version of DMS are able to take as inputs the red and NIR bands of MODIS separately. However, this approach has not been tested in this study.

\item The learned parameters (slope and intercept for TsHARP and ATPRK, and decision trees for DMS) at low resolution are considered scale-invariant and used at fine resolution to provide a first estimate of $T^{(h)}_{sr}$.

\begin{equation}
    T'^{(h)}_{sr} = f(V^{(h)}_{obs})
\end{equation}

\item Finally, a small scale residual correction is done:

\begin{equation} \label{eq:5}
    T^{(h)}_{sr} = T'^{(h)}_{sr} + \Delta^{(h)} T
\end{equation}

\noindent The definition of the small scale residuals, $\Delta^{(h)} T$, is different depending on the method. In the case of TsHARP and DMS $\Delta^{(h)} T$ is obtained by a nearest-neighbor interpolation of $\Delta^{(l)} T$. In the case of ATPRK, $\Delta^{(h)} T$ is obtained by kriging from $\Delta^{(l)} T$~\cite{GraneroBelinchon2019}.
\end{itemize}

A detailed description of TsHARP and ATPRK can be found in~\cite{GraneroBelinchon2019} and references therein, while a detailed description of DMS can be found in~\cite{Gao2012}. The TsHARP and ATPRK implementations used for benchmarking were implemented following exactly~\cite{GraneroBelinchon2019} and are available at https://github.com/cgranerob/ThUnmpy. The DMS approach used as a benchmark was developped by Guzinski and Nieto~\cite{Guzinski2019}, and is freely accessible at https://github.com/radosuav/pyDMS.

\subsubsection{U-Net trained under scale invariance hypothesis}

We train a Neural Network (NN) model with the same architecture as SIF-CNN-SR, described in section~\ref{sec:arch}, and trained on the same MODIS dataset, described in section~\ref{sec:datatrain}, but with a reduced scale training approach. We call it SC-Unet. This model is trained in a supervised way on MODIS data at degraded spatial resolutions~\cite{Nguyen2022}, \textit{i.e.} the LST of MODIS at 1 km of resolution is used as reference while the inputs are degraded versions of the LST and the NDVI at 4 km and 1 km of spatial resolution, respectively. Thus, SC-Unet is trained to provide LST at 1 km from LST and NDVI at 4 km and 1 km respectively. The loss function used to train the model is just the MSE (Mean-Squared Error) between the LST provided by the model and the MODIS LST at 1 km of resolution. Once the model is trained, it is used to provide LST at 250 m from LST at 1 km and NDVI at 250 m. Thus, SC-Unet grounds on the scale-invariance hypothesis. The model was trained during 100 epochs, with batch size 32 and learning rate 0.0001.

\subsection{Evaluation Metrics}\label{sec:eval}

Seven complementary metrics were used to measure the performances of super-resolution models: Root Mean Squared Error (RMSE), Root Mean Square Error averaged only on the quartile of pixels with higher gradients (RMSE$_{75-100}$)~\cite{Rama2022}, Structural Similarity Index Measure (SSIM)~\cite{Wang2004}, Learned Perceptual Image Patch Similarity (LPIPS)~\cite{Zhang2018}, Frequency Restoration Rate (FRR)~\cite{Michel2024}, Frequency Restoration Overshoot (FRO)~\cite{Michel2024} and the RMSE between attenuation spectra $\mathbf{F}(\nu)$, (RMSE($\mathbf{F}(\nu)$). All of them are based on comparisons between super-resolution LST ($T^{(h)}_{sr}$) and the reference LST ($T^{(h)}_{ref}$). While the RMSEs and SSIM are comparisons performed directly in the LST space, LPIPS performs a comparison in a learned latent space and FRR, FRO and RMSE($\mathbf{F}(\nu)$) are evaluations performed in the Fourier spectral domain.

\subsubsection{Metrics on the space of LST}

The RMSE measures the overall error between $T^{(h)}_{sr}$ and $T^{(h)}_{ref}$:

\begin{equation}
RMSE(T^{(h)}_{sr},T^{(h)}_{ref}) = \sqrt{\frac{1}{N}||(T^{(h)}_{sr} - T^{(h)}_{ref}||^2_F}
\end{equation}

\noindent where $N$ is the number of pixels of the images and $||\,||^2_F$ is the Frobenius norm.
 
%\begin{equation}
%RMSE(T^{(h)}_{sr},T^{(h)}_{ref}) = \sqrt{\frac{1}{X\,Y}\sum_{x=1}^{X}\sum_{y=1}^{Y}(T^{(h)}_{sr}(x,y) - T^{(h)}_{ref}(x,y))^2}
%\end{equation}

%\noindent where $X$ and $Y$ are the number of pixels of the image along each spatial dimension and $T^{(h)}_{sr}(x,y)$ and $T^{(h)}_{ref}(x,y)$ the LST values at the coordinates $x$ and $y$ of the pixel.

We also study the RMSE on specific areas of the images defined in function of the amplitude of the gradients of their pixels. The RMSE$_{75-100}$ is estimated on pixels with an amplitude of the gradient in the fourth quartile, \textit{i.e.} on areas with very heterogeneous textures, where super-resolution is more difficult.

The SSIM measures the perception-based similarity between two images $x$ and $y$ and between their variations~\cite{Wang2004}. We use an implementation based on the computation of the mean and variance within a sliding square window of size $7$ pixels ($1750$ m). This provides a SSIM value per pixel of the image. Finally, the SSIMs are averaged to get an overall score per image. The SSIM between two windows, one for each image being compared, both centered on the pixel $(x_1,x_2)$ is: 

\begin{equation}
SSIM(x,y) = l(x,y)^{\beta_1} c(x,y)^{\beta_2} s(x,y)^{\beta_3}
\end{equation}

\noindent where $l$, $c$ and $s$ are the radiance, contrast and structure terms and are defined as:

\begin{align}
l(x,y) &= \frac{2\mu_x\mu_y + c_1}{\mu_x^2 + \mu_y^2 + c_1} \\
c(x,y) &= \frac{2\sigma_x\sigma_y + c_2}{\sigma_x^2 + \sigma_y^2 + c_2} \\
s(x,y) &= \frac{\sigma_{xy} + c_3}{\sigma_x\sigma_y + c_3}
\end{align}

\noindent where $\mu_x$ and $\mu_y$ are, respectively, the mean value of the pixels in the square window for each image, $\sigma_x$, $\sigma_y$ and $\sigma_{xy}$ are respectively the variance of the pixels in the window of each image and the cross covariance. $c_1 = (K_1 L)^2$, $c_2 = (K_2 L)^2$ and $c_3 = c_2 / 2$ are constants that avoid the denominator getting reduced to $0$. $K_1 = 0.01$, $K_2 = 0.03$ and $L$ is the data range corresponding to the difference between the maximum and the minimum LST in both images. $\beta_1$, $\beta_2$ and $\beta_3$ are set to 1, as usually. Images that are close in both intensity and variations present SSIM values close to $1$.

\subsubsection{Metrics on latent and Fourier spaces}

LPIPS is a deep similarity metric defined by Zhang et al.~\cite{Zhang2018}. LPIPS combines the Mean Squared Error between several latent representations of the super-resolution image and the reference. These representations are obtained using the intermediary features of a pretrained CNN such as VGG~\cite{He2016}. The greater the resemblance between the two images, the lower LPIPS will be.

Three metrics were used to compare directly the frequency content of $T^{(h)}_{sr}$ and $T^{(h)}_{ref}$ in the two-dimensional Fourier space as proposed in~\cite{Michel2024}. To facilitate the comparison, isotropy is considered and then an average is done on the angular dimension of the two-dimensional Fourier spectra, \textit{i.e.} only the radial dimension of the spectra is considered. 

The power spectra are computed following: 

\begin{equation}
F(\nu) = \frac{1}{\# U_{f_m,f_M}}\sum_{(\nu_1,\nu_2) \in U_{f_m,f_M}} |F(\nu_1,\nu_2)|
\end{equation}

\noindent where $U_{f_m,f_M}=\left\lbrace (\nu_1, \nu_2) : f_m \leq \sqrt{\nu_1^2 + \nu_2^2} < f_M \right\rbrace$ defines the discrete spatial frequencies lying in the ring defined by $f_m$ and $f_M$ and $\#U_{f_m,f_M}$ is the number of elements in $U_{f_m,f_M}$.

The attenuation spectra are:

\begin{equation}
\textbf{F}(\nu) = 10(log_{10}(F(\nu)) - log_{10}(F(0)))
\end{equation}

The three metrics used are: 

\begin{itemize}
    \item the RMSE($\mathbf{F}(\nu)$) between the attenuation spectrum of ASTER LST $T^{(h)}_{ref}$ and each super-resolution product $T^{(h)}_{sr}$.
    
    \item the FRR, described by Michel et al.~\cite{Michel2024}, that compares the attenuation spectra of: the bicubic interpolation of $T^{(l)}_{obs}$ considered here as the worst possible super-resolution methods, the reference LST $T^{(h)}_{ref}$, and the super-resolution products $T^{(h)}_{sr}$. It measures the improvement in frequency reconstruction of $T^{(h)}_{sr}$ compared to bicubic interpolation of $T^{(l)}_{obs}$. %This metric is computed as the difference between the energy loss of the bicubic interpolation compared to the ground truth and the energy loss of $T^{(h)}_{sr}$ compared to the ground truth.
    \item the FRO, described by Michel et al.~\cite{Michel2024}, measuring the overshoot of the attenuation spectrum of $T^{(h)}_{sr}$ compared to $T^{(h)}_{ref}$.
    %\item the FRU measuring the loss of energy in frequencies between $T^{(h)}_{sr}$ and the bicubic interpolation of $T^{(l)}_{obs}$.
\end{itemize}

\section{Results}

\subsection{Performance of the models in central Europe}

Table~\ref{tab:1} shows the performance of Bicubic interpolation, TsHARP, ATPRK, DMS, SC-Unet, SIF-CNN-SR1 and SIF-CNN-SR2 evaluated with the seven metrics presented in section~\ref{sec:eval} averaged over the full evaluation dataset of central Europe. The best RMSE is obtained with SC-Unet and SIF-CNN-SR2. Indeed, these two models are the only ones with a RMSE value smaller than $2$ K. However, only a small difference of 0.2 K appears between these models and bicubic interpolation and only $0.4$ K with the highest RMSE value for ATPRK with overlapping standard deviation values between all the methods. Nevertheless, these metrics are averaged on the whole dataset and we expect the standard deviation of the RMSE will decrease with the size of the dataset. In addition, the RMSE for each method can fluctuate from one image to another, meaning that the most efficient method can change according to the image. SC-Unet and SIF-CNN-SR2 show the best performances on $RMSE_{75-100}$, \textit{i.e.} when only highly heterogeneous areas are considered in the RMSE estimation. SC-Unet, SIF-CNN-SR1 and SIF-CNN-SR2 present the highest SSIM values with very similar performances: SC-Unet has SSIM=0.64, SIF-CNN-SR1 has SSIM=0.61 and SIF-CNN-SR2 has SSIM=0.62. Concerning the metrics focusing on textures, the best LPIPs score is obtained with SIF-CNN-SR1 (LPIPS=0.28) closely followed by SC-Unet (LPIPS=0.29). In the Fourier domain, the model SIF-CNN-SR1 has the smallest RMSE($\mathbf{F}(\nu)$), indicating that the attenuation spectra obtained with this model is the closer to the ASTER spectra. In addition, it presents the best FRR and so the best improvement of frequency reconstruction compared to the bicubic interpolation. Finally, the best FRO is obtained with SC-Unet and bicubic. Very importantly FRO and FRR should be analyzed together since while FRO is blind to underestimation of the attenuation spectra, FRR is blind to overestimation. On the other hand the analysis of RMSE($\mathbf{F}(\nu)$) should be done with care since this metric does not discriminate between under and overestimation of the spectrum.

Table~\ref{tab:1} allows us to conclude that SC-Unet and SIF-CNN-SR2 present the smaller overall errors on RMSE. However, SIF-CNN-SR1 seems to better recover the textures as evaluated with LPIPS. In the Fourier domain, which is indicative both of textures and overall errors, the best performances are obtained by the proposed model SIF-CNN-SR1. SIF-CNN-SR1 generates LSTs with more high-frequency content compared to the other models as evaluated by FRR and RMSE($\mathbf{F}(\nu)$) but it slightly overestimate textures as indicated by FRO. These results quantify the presence of more visible textures inside the super-resolution made by SIF-CNN-SR1. The three convolutional NN outperform state of the art approaches by a high margin in most metrics. %\com{c'est enbetant qu'un modele NN avec une hypothèse de scale invariance fonctionne aussi bien que les 2 autres, à part la haute frequence qu'il faut mettre en avant}

\begin{table*}
 \caption{Average RMSE, RMSE$_{75-100}$, SSIM, LPIPS, FRR, FRO and RMSE($\mathbf{F}(\nu)$) over the evaluation dataset containing 79 daytime images over central Europe for the five studied super-resolution approaches. Bold indicates the best average performance for each given metric and brackets () indicate the standard deviation.}\label{tab:1}
 \begin{center}
\begin{tabular}{ |p{2cm}|p{1.5cm}|p{1.5cm}|p{1.5cm}|p{1.5cm}|p{1.5cm}|p{1.5cm}|p{1.5cm}|p{1.5cm}|}
 \hline
 Methods& RMSE          & RMSE$_{75-100}$ & SSIM       & LPIPS    & FRR  & FRO  &RMSE($\mathbf{F}(\nu)$) \\
 \hline
 Bicubic& 2.1 (0.7)     & 2.9 (0.8)            & 0.46 (0.09)          & 0.39 (0.03)          & 0.00 (0.00) & \textbf{0.00 (0.00)} &  5.0 (0.9) \\
 TsHARP & 2.2 (0.7)     &  2.8 (0.9)           & 0.56 (0.08)         & 0.34 (0.05)          & 0.95 (0.05) &  0.04 (0.01) & 2.9 (0.6) \\
 ATPRK  & 2.3 (0.7)     & 2.8 (0.9)           & 0.54 (0.08)         & 0.32 (0.04)          & 0.95 (0.10) &  0.04 (0.01) & 2.6 (0.7) \\
 DMS    & 2.1 (0.7)     & \textbf{2.5 (0.9)}   & 0.58 (0.09) & 0.31 (0.03)          & 0.96 (0.06) & 0.04 (0.01) & 2.6 (0.6) \\
 SC-Unet    & \textbf{1.9 (0.7)} & \textbf{2.5 (0.9)}            & \textbf{0.64 (0.07)}          & 0.29 (0.04)         & 0.62 (0.13)  & \textbf{0.00 (0.00)} & 1.8 (0.6) \\
 SIF-CNN-SR1   & 2.2 (0.7)     & 2.6 (0.8)           & 0.62 (0.08) & \textbf{0.28 (0.04)} & \textbf{0.98 (0.03)} & 0.03 (0.02) &\textbf{1.6 (0.7)} \\
 SIF-CNN-SR2   & \textbf{1.9 (0.7)} & \textbf{2.5 (0.8)}            & 0.61 (0.07)         & 0.32 (0.04)          & 0.79 (0.10) & 0.01 (0.01) & 1.8 (0.3) \\
 \hline
\end{tabular} 
\end{center}
\end{table*}

\begin{figure}[h!]
 \includegraphics[width=\linewidth]{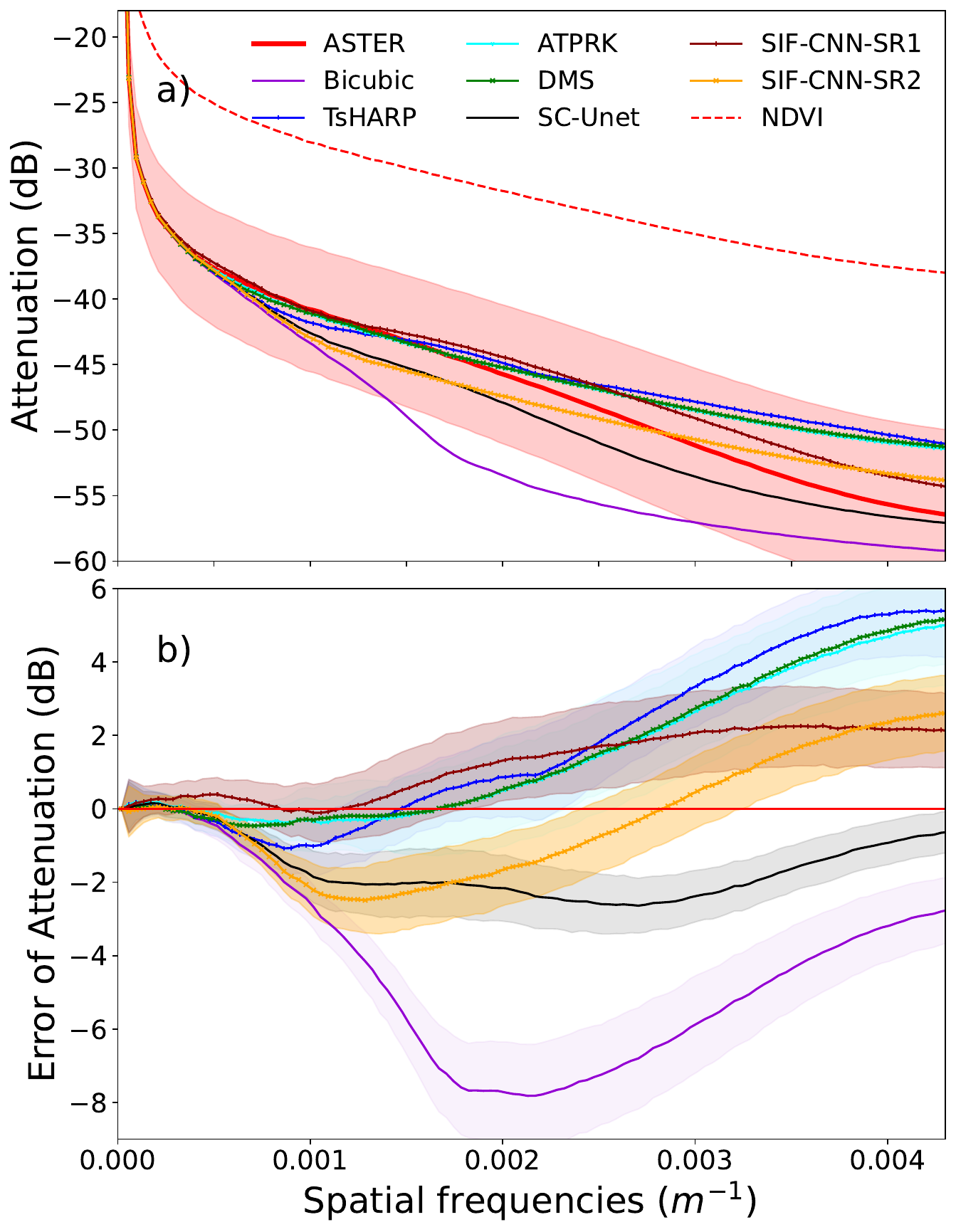}
 \caption{a) Attenuation spectra of the ASTER LST (red), LST obtained with statistical super-resolution methods TsHARP and ATPRK (blue), DMS (green), SC-Unet (black), SIF-CNN-SR1 (brown) and SIF-CNN-SR2 (orange). The attenuation spectra of the MODIS NDVI is also shown in dashed red. The curves correspond to the mean spectra over the full test dataset. The shadow around the ASTER attenuation spectra correspond to $\pm$ a standard deviation around the mean value. b) Mean Error between ASTER attenuation spectrum and the attenuation spectra of TsHARP and ATPRK (blue), DMS (green), SC-Unet (black), SIF-CNN-SR1 (brown) and SIF-CNN-SR2 (orange). The curves correspond to the mean error over the full test dataset and the shadowed areas correspond to one standard deviation around the mean. The red horizontal line indicates the zero.} \label{fig:avgspectra}
\end{figure}

Figure~\ref{fig:avgspectra} a) shows the attenuation spectra of ASTER LST, the different super-resolution LST products and the MODIS NDVI averaged over the 79 images of the evaluation dataset. We can observe that the bicubic interpolation and SC-Unet tend to underestimate the middle range and small scale textures of the LST while TsHARP, ATPRK, DMS and SIF-CNN-SR1 tend to overestimate them. SIF-CNN-SR2 underestimates the middle range scales while overestimating small scales. TsHARP, ATPRK, DMS and SIF-CNN-SR2 present an attenuation spectra following the shape of the attenuation spectra of the MODIS NDVI. This can be understood since the MODIS NDVI is used in these models to define the small scale textures. Very interestingly, the attenuation spectra of SIF-CNN-SR1 and SC-Unet present a dynamic across scales that follows the one of ASTER LST, even if they also use MODIS NDVI to define the small scales. SIF-CNN-SR1 closely matches the ASTER LST spectra until scales $\sim 500$ m and overestimates the texture for smaller scales. On the other hand, SC-Unet underestimates the textures for scales smaller than $1$ km.
Figure~\ref{fig:avgspectra} b) shows the error between the ASTER attenuation spectrum and the attenuation spectra of each super-resolution product. SIF-CNN-SR1 avoids significant errors at low and intermediary resolutions compared to SC-Unet and SIF-CNN-SR2 and it also avoids significant errors at high resolutions compared to DMS, TsHARP and ATPRK. 

\begin{figure*}[ht]
 \includegraphics[width=\linewidth]{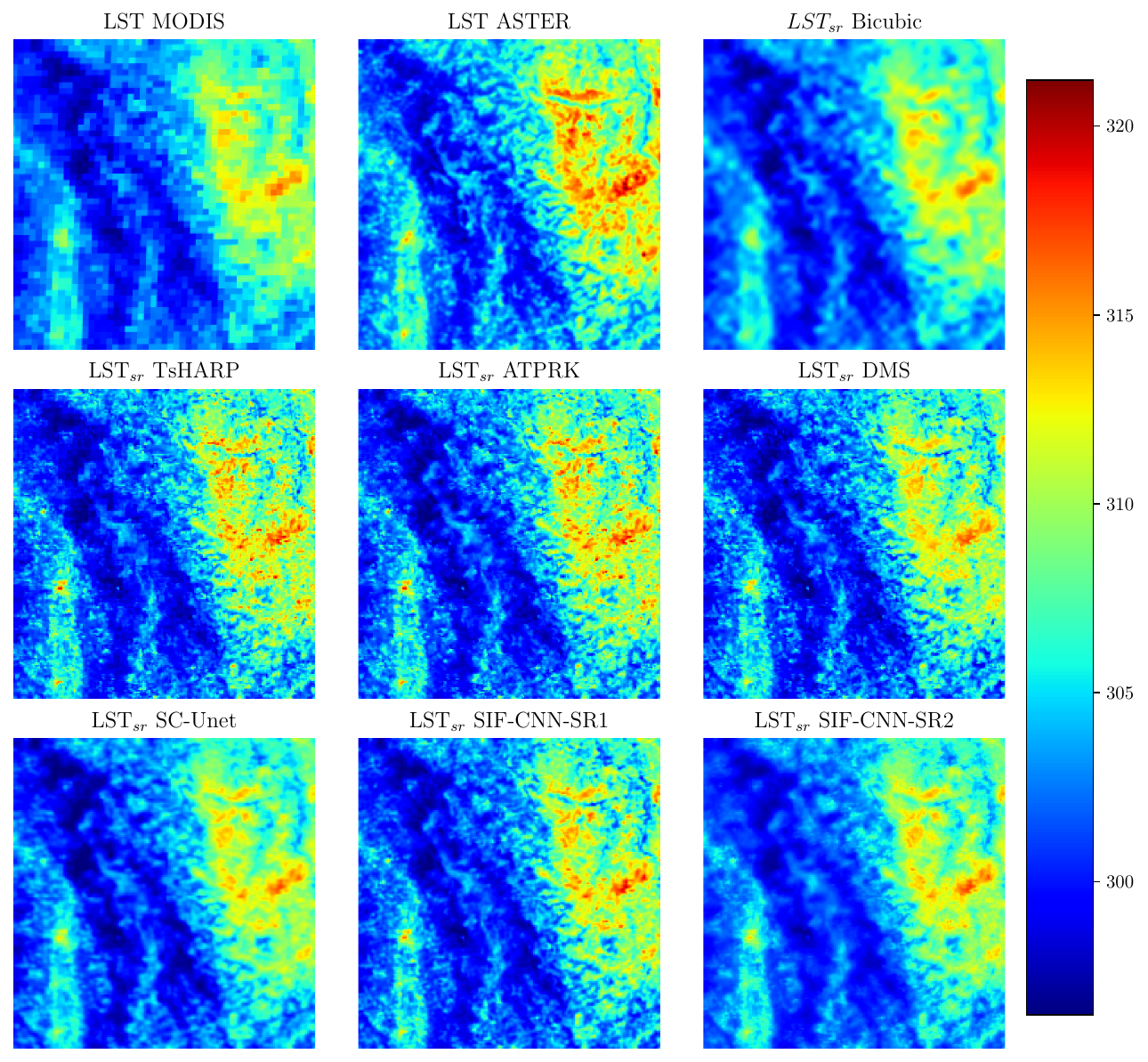}
 \centering
 \caption{For one random image from the validation dataset, visualization of the LST of MODIS and ASTER respectively at 1km and 250m of spatial resolution alongside the super-resolution LST obtained with the different approaches.}\label{fig:vis0}
\end{figure*}

For a qualitative and deeper statistical analysis of the models' performances, we randomly choose one image from the evaluation dataset, then we study it visually as well as its attenuation spectrum and its distribution of the values of the high-pass filtered LST\footnote{The same analysis for each image of the validation dataset can be found in the supplementary materials and at https://github.com/cgranerob/Land-Surface-Temperature-Super-Resolution-with-a-Scale-Invariance-Free-Neural-Approach.}. Figure~\ref{fig:vis0} allows the visual comparison of the observed MODIS LST at 1 km and the LST from ASTER at 250 m together with the LSTs at 250 m obtained with the different methods explained in section~\ref{sec:benchmarks}.
While Bicubic is significantly blurred compared to ASTER LST and is visually similar to the MODIS at 1 km, TsHARP and ATPRK present similar patterns with more textures than ASTER LST. SC-Unet and SIF-CNN-SR2 also present blurrier LSTs than ASTER and the methods that seem to better recover the LST from ASTER are DMS and SIF-CNN-SR1. %For the Figure \ref{fig:vis1}, we observe a clear different texture between MODIS and ASTER at the bottom left corner of the image where hot spots of ASTER LSTs are more spatially contrasted with colder spots than MODIS LST, confirming this pre-existing bias between both sensors. As a result, all the methods overestimate the LST over this area. For SC-Unet, SIF-CNN-SR2 and Bicubic, the same observations can be made than for the first image as they are the blurriest with the two former having more similar textures with ASTER LST. Nevertheless, there is a difference with the first image because DMS presents slightly less similar patterns than ATPRK and is similar to Tsharp. Thus, the methods that seem to better recover the LST from ASTER are ATPRK and SIF-CNN-SR1 instead of DMS and SIF-CNN-SR1 for the first imafe. For the Figure \ref{fig:vis2} showing the visual comparison between ASTER LST, MODIS LST and the super-resolved LSTs from all the methods for the third selected image, the same observations are made. Bicubic, SC-Unet and SIF-CNN-SR2 are the blurriest and Tsharp and for ATPRK, DMS and SIF-CNN-SR1 present more textures similar to ASTER LST. DMS seems to better recover the LST in this case. Last, all the methods also overestimate the LST at the bottom left corner and we observe the bias between ASTER LST and MODIS LST. }

\begin{figure}[h!]
 \includegraphics[width=\linewidth]{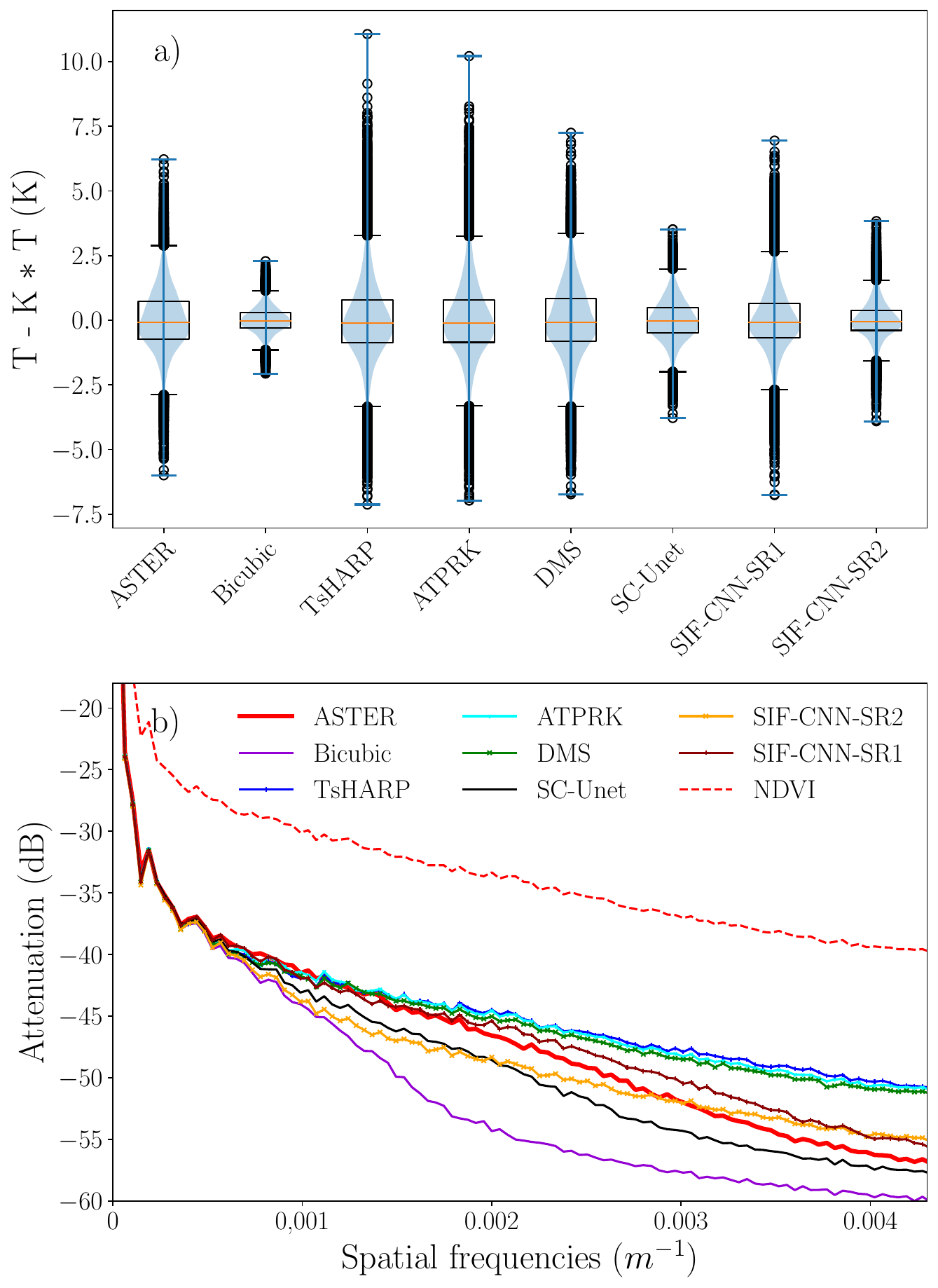}
 \caption{Statistical analysis of the image visualized in figure~\ref{fig:vis0}. a) Boxplots and Violinplots representing the statistical distribution of the values of the high pass filtered LST (see equation~\ref{eq:highpass}) of ASTER and obtained with the different super-resolution approaches. b) Attenuation spectra of the ASTER LST (red), LST obtained with statistical super-resolution methods TsHARP and ATPRK (blue), DMS (green), SC-Unet (black), SIF-CNN-SR1 (brown) and SIF-CNN-SR2 (orange). The attenuation spectra of the MODIS NDVI is also shown in dashed red.}\label{fig:violon0}
\end{figure}

Figure~\ref{fig:violon0} a) represents the high-frequency filtered content of the super-resolved LSTs. ATPRK and TsHARP tend to overestimate the textures, while Bicubic, SC-Unet and SIF-CNN-SR2 tend to underestimate them. DMS and SIF-CNN-SR1 present the closest distributions of high-frequency filtered LSTs. Figure~\ref{fig:violon0} b) illustrates the attenuation spectrum of ASTER LST in red, MODIS NDVI in dashed red and super-resolution LST from bicubic interpolation in magenta, TsHARP and ATPRK in blues and DMS in green, SC-Unet in black, SIF-CNN-SR2 in orange and SIF-CNN-SR1 in brown. As expected from previous results, bicubic interpolation SC-Unet and SIF-CNN-SR2 tends to underestimate mid-range and high frequencies of the LST. On the other hand, ATPRK, TsHARP, and DMS tend to overestimate high frequencies of the LST. SIF-CNN-SR1 presents the attenuation spectra closer to ASTER but with slightly overestimated high frequencies.

%\com{For the attenuation spectra on Figure \ref{fig:att1}, SC-Unet is the closest to the ASTER attenuation spectrum, SIF-CNN-SR2 and SIF-CNN-SR2 follow it, SIF-CNN-SR2 overestimates at high spatial frequencies and SIF-CNN-SR1 overestimates until high frequencies where it is similar to SIF-CNN-SR2, which is also observed on Figure \ref{fig:att}.} 

\section{Discussion}\label{sec:discussion}

\subsection{Comparison of the different models}

All the super-resolution models studied in this work use as principal hypothesis to extract high resolution textures the relationship existing between the LST and the NDVI. This relationship has been demonstrated by many \cite{Cai2018, Govil2019, Kumar2015}. However, this hypothesis is exploited differently by the different models.

First, TsHARP, ATPRK, DMS and SC-Unet use a scale-invariance hypothesis, \textit{i.e.} the statistical models are learned at a spatial resolution that is lower than the spatial resolution of application. In the case of SIF-CNN-SR1 and SIF-CNN-SR2 there is not the scale-invariance hypothesis since the training is performed directly at the spatial resolution of application. 
%
%SC-Unet, $f_{\theta}$ has been trained using LST and NDVI at degraded resolutions, and so SC-Unet uses a scale-invariance hypothesis. Compared to the similar work of~\cite{Nguyen2022}, that inspired the architecture of our models, SC-Unet performs better because it uses the NDVI in the learning process, adding a constraint to the minimization problem that allows to find a better solution. Nevertheless, SC-Unet is not as performant as SIF-CNN-SR1 to recover the high-frequency content.

Second, TsHARP, ATPRK and DMS can be considered as models that predict the high resolution LST only from the high resolution NDVI: $T^{(h)}_{sr}=f_{\theta}(V^{(h)}_{obs})$ where $f_{\theta}$ has been trained using $(T^{l}_{obs},V^{l}_{obs})$. In addition, they use a residual correction to overcome the use of an analytical relationship between NDVI and LST. This correction is computed at low resolution and interpolated at high resolution through a nearest neighbor approach for TsHarp and DMS or kriging for ATPRK, see \ref{sec:benchmarks}. On the other hand, SC-Unet, SIF-CNN-SR1 and SIF-CNN-SR2 predict the high resolution LST by directly taking as input both NDVI and LST. Our CNN models do not need any a posteriori correction.

Third, classical statistical methods require to learn a new $f$ for each new data to process. These methods mostly learn to predict LST from the input NDVI through the function $f$. As NDVI does not contain important instantaneous drivers of the LST such as meteorological conditions, this relationship can only hold for a given instant, and can not generalize to different locations or times. This leads to a learned $f$ that specifically targets the observed area, but it is also a limitation that prevents to use more complex mapping $f$, because of the cost of re-training and the limited training data available to do it. On the other end, our proposed approach predict high resolution LST from both low resolution LST and high resolution NDVI, and is thus able to generalize to other locations and times, with comparable LST patterns, by using a more complex mapping $f$ based on CNN, while saving inference time due to the absence of re-training.

Summarizing, the three main differences between the presented models are 1) the scale-invariance hypothesis, 2) training a function that explicitly depends on both NDVI and LST and 3) training a function able to generalize to different locations and times. These three differences can explain the better performance of our models with respect to the state of the art.

Another criteria to discriminate the super-resolution algorithm is their inference time since it strongly variates from one algorithm to another. The mean times for inference over the full test dataset are Bicubic 0.00007 s, TsHARP 0.01 s, ATPRK 19.56 s, DMS 0.31 s, and the three NNs  0.03 s. Our NNs present inference times of the order of TsHARP, ten times smaller than DMS and more than one hundred times smaller than ATPRK.

\subsection{SIF-CNN-SR1 \textit{versus} SIF-CNN-SR2}

The main difference between SIF-CNN-SR1 and SIF-CNN-SR2 is the definition of the texture operator $G$. While in SIF-CNN-SR1, $G$ corresponds to the gradients defined with Sobel filters, in SIF-CNN-SR2, $G$ is a high-pass filter defined through a convolution with a Gaussian kernel. The gradient allows SIF-CNN-SR1 to generate LST products whose attenuation spectra better follows the evolution across spatial frequencies of the attenuation spectra of ASTER. The high-pass filter seems to not characterize scales between $1$km and $300$m leading to underestimated attenuation spectra of SIF-CNN-SR2 at these scales.

In particular, SIF-CNN-SR1 is promising for applications on LST from other sensors such as TRISHNA since it presents performance at the level of the state of the art, or even better when we look at the Fourier space metrics. Indeed, at higher spatial resolutions ($<100$ m) the scale-invariance hypothesis seems to be more problematic~\cite{GraneroBelinchon2019}. The size of the pixel starts to be of the order of the observed objects and the heterogeneity of some landscapes make the scale-invariance hypothesis non-adapted. In addition, the scale-invariance hypothesis is specially not adapted for complex models such as Deep Learning models or DMS that fit better the NDVI-LST relationship at the scale of training. We expect models based on simple linear regression to be less impacted by the scale-invariance hypothesis~\cite{DelairRAQRS2024}, \textit{i.e.} simple models define more generic LST-NDVI relationships, and so even if the learned relationship is not scale-invariant, the impact of this hypothesis on final performance is reduced.

\subsection{On the evaluation metrics}

In this study, we used seven complementary evaluation metrics. On the one hand, we studied metrics estimated directly on LST values: 1) RMSEs on the full images and in some parts of the images where the gradients are especially important and 2) SSIM. The RMSEs look at the average difference in Kelvin between super-resolution and reference images. Even if the information provided by these metrics is interesting, the RMSE, looking directly at LST values (not textures) and being averaged over whole areas, is not necessarily adapted to evaluate super-resolution performance. The SSIM looks at differences in the perceptual structure of the images. It depends on the LST values of the image and on their spatial distribution. SSIM is rarely discriminant in super-resolution. SSIM and RMSE may be prone to biases in the presence of radiometric or geometric distortions in cross-sensors dataset, therefore they will be jointly analyzed with complementary metrics~\cite{Michel2024}. On the other hand, we studied metrics focusing on the characterization of textures: 1) LPIPS working on latent learned representations and 2) RMSE($\mathbf{F}(\nu)$), FRR and FRO that are metrics on the Fourier domain. These metrics are less sensitive to specific LST values and are more concerned by the scales resolved by the models. We consider that they are more representative for the characterization of the performance of super-resolution models. To the best of our knowledge, these latter metrics are not commonly used in studies of super-resolution of LST. This work highlights the interest of considering them for future studies.

\subsection{On the relationship between the NDVI and the LST}

The models considered in this study rely on an inverse relationship between the NDVI and the LST. In fact, the NDVI is associated with vegetation cover, and, at daytime, LST tends to be cooler for vegetated areas than for bare soils or impervious surfaces, so the LST and the NDVI are generally negatively correlated. Several works have studied the relationship between LST and NDVI in detail illustrating some variability depending on time, land cover and sensor. \cite{Sun2007} showed a positive relationship between the LST and the NDVI during winter with GOES-8 data at 8km spatial resolution. \cite{Guha2020} found varying negative correlations according to the season with low negative correlations during winter with Landsat 8 data. They also found that this relationship varies according to land cover, with the highest negative correlations for vegetated areas as expected and lower for built-up areas. For water bodies, the LST can be cooler than for vegetation but the NDVI generally has negative or close to zero values as pointed by~\cite{Cai2018}.

The models SIF-CNN-SR1 and SIF-CNN-SR2 use a negative $\gamma$ value due to the general assumption of the inverse relationship between the NDVI and the LST. However, these models are trained considering all seasons and different land cover types, and this can degrade their performance on some specific images with important water bodies. As a perspective, an adaptive $\gamma$ value could be defined for each image to consider specific cases due to the variability of this relationship. %The inverse relationship between LST and NDVI is also at the core of classical super-resolution methods as stated in~\ref{sec:relatedworks}.

\subsection{On the use of ASTER for validation}

Another contribution of this work is the generation of a test database with pairs of concomitant ASTER LST and MODIS (LST,NDVI) images that will be useful for future studies on LST super-resolution\footnote{This database is available at https://github.com/cgranerob/Land-Surface-Temperature-Super-Resolution-with-a-Scale-Invariance-Free-Neural-Approach.}. The ASTER and MODIS LST products are obtained with the TES algorithm in order to reduce differences in the LST retrieval. However, we observed a bias between these LSTs. This bias can be explained since the number and spectral range of the TIR bands of ASTER and MODIS are different~\cite{Hulley2012}. In addition, it may depend on the spatial heterogeneity of the observed landscape~\cite{Liu2006, Liu2009}. Also, ASTER captures the LST variability at 90m which allows to better recover fine textures when MODIS has a smoother LST due to its spatial resolution of 1 km. These differences, together with other parameters such as the viewing angle, the intrinsic error of the TES algorithm or co-registration errors between ASTER and MODIS images, lead to a bias between the ASTER and MODIS LST that is different for the different couples of images of the validation dataset. Most of the images present a bias between 0 K and 2 K and only two couples present a bias higher than 2 K. The evaluation metrics based on the RMSE depend linearly on this bias: the higher the bias, the higher the RMSE. This is not the case for LPIPS, SSIM or Fourier space metrics which emphasize the interest of these metrics for the evaluation of super-resolution methods. Although an attempt to reduce the bias between ASTER and MODIS was found~\cite{Liu2007}, this is an ongoing issue that should be treated in the future to improve the validation of super-resolution methods~\cite{Yoo2020}. 

The ASTER LST images at 250 m of spatial resolution used for validation are obtained by low-pass filtering the initial ASTER LST at 90 m, see section~\ref{sec:evaldata}. This low-pass filtering introduce a blurring effect that depends on the filter but its impact is supposed insignificant for spatial scales larger than 250 m. The energy content, as characterized by the attenuation spectra, of MODIS NDVI at small scales $\sim250$m is significantly higher than the energy content of ASTER LST at these scales. This implies that the MODIS NDVI presents more small scale textures, \textit{i.e.} high-frequency content, than the ASTER LST. This difference is in part explained by the fact that NDVI and LST are different physical variables, and so, the NDVI and ASTER textures would not be identical. However, the low-pass filter used to provide ASTER LST at 250 m could also attenuate small scales. Unfortunately, these two effects are completely mixed and cannot be discriminated. These remarks need to be considered when using the test database for future studies.

\section{Conclusions and perspectives}

In order to overcome the trade-off between temporal and spatial resolution of the TIR sensors, current stat-of-the-art methods such as Tsharp, ATPRK or DMS rely on a scale-invariance hypothesis to perform the super-resolution of LST images: models optimized at coarse scales are applied at small ones. However, this hypothesis can lead to an important performance loss. In this work, we proposed a Scale-Invariance-Free approach to train two Convolutional Neural Networks, SIF-CNN-SR1 and SIF-CNN-SR2, that increase the resolution of MODIS LST images from 1 km to 250 m. We also propose a CNN sharing the same architecture as SIF-CNN-SR but being trained under the scale-invariance hypothesis, SC-Unet. This model allows to compare the proposed scale-invariance-free training approach with a classical reduced scale one, described in section~\ref{sec:benchmarks}. All the models studied in this work use the MODIS NDVI at 250 m to provide information at high spatial resolution.

To evaluate the performance of each super-resolution method, we have generated a validation dataset by pairing concomitant LST images from ASTER with couples of (NDVI-LST) images from MODIS. Both ASTER LST and MODIS NDVI are provided at 250 m of spatial resolution while MODIS LST is provided at 1 km of resolution. Consequently, we generated an analysis-ready-dataset of MODIS and ASTER concomitant images that can be used for benchmarking super-resolution methods.

We compared the three proposed deep learning models with the aforementioned state-of-the-art methods. The results show that SIF-CNN-SR1 outperforms the other methods. Also, we showed that RMSE and SSIM are partially adapted for the evaluation of super-resolution methods since they do not focus on the characterization of textures. Indeed, metrics focusing on textures such as LPIPS and Fourier-space metrics appear as more appropriated for this case of studies.

Future research directions include: considering nighttime LST images without any concomitant NDVI image or testing the generalization of the model for different geographical regions. Also, the choice of an adaptive parameter $\gamma$ to capture the variability of the relationship between the NDVI and the LST could lead to better performances. Finally, the ability of the models to operate at higher spatial resolutions should be investigated as future thermal missions are planned such as LSTM, TRISHNA or SBG that will provide LST images at around 50 m. Indeed, the monitoring of the LST for a wide range of applications would benefit from time series of LST images at both high temporal and spatial resolutions.

\bibliographystyle{IEEEtran} 
\bibliography{bibliography}

\end{document}